\documentclass{article} 
\usepackage[final]{Template-2026/colm2026_conference}

\usepackage{microtype}
\usepackage{hyperref}
\usepackage{url}
\usepackage{booktabs}
\usepackage{amsmath,amssymb}
\usepackage{multirow}
\usepackage{graphicx}
\usepackage{xcolor}
\usepackage{subcaption}
\usepackage{algorithm}
\usepackage{algpseudocode}
\usepackage{tabularx}
\usepackage{float}
\floatstyle{ruled}
\restylefloat{algorithm}

\usepackage{lineno}

\definecolor{darkblue}{rgb}{0, 0, 0.5}
\hypersetup{colorlinks=true, citecolor=darkblue, linkcolor=darkblue, urlcolor=darkblue}

\title{CHASE: How Content Ecosystems Are Reshaped When Ranking Is the Only Target}

\author{
Qianwen Gao\textsuperscript{1}\thanks{Equal contribution.},
Zichang Su\textsuperscript{2}\footnotemark[1],
Yiwen Hou\textsuperscript{1},
Arlen Kumar\textsuperscript{1} \&
Leanid Palkhouski\textsuperscript{1}
\\
\textsuperscript{1}University of California, Berkeley
\qquad
\textsuperscript{2}Zhejiang University
\\
\texttt{\{kaiagao,leanid\}@berkeley.edu}
}

\newcommand{\Dt}{\mathcal{D}_t}
\newcommand{\phivec}{\boldsymbol{\phi}}

\begin{document}

\ifcolmfinal
\fi

\maketitle
\pagestyle{fancy}
\lhead{Published as a conference paper at COLM 2026}
\begin{abstract}
Generative Engine Optimization (GEO) is increasingly used to improve content
visibility in LLM-based retrieval systems, yet its population-level effects
under repeated optimization remain poorly understood.
We introduce Content Homogenization under rAnking Signal Exploitation
(CHASE)\footnote{\url{https://github.com/kaiagaoo/CHASE}},
a controlled simulation framework for studying how content ecosystems are reshaped
when creators repeatedly adapt documents to an LLM ranking signal.
We use ranking as a proxy for source visibility and validate this abstraction
against citations in grounded generated responses, obtaining a rank--citation
AUC of $0.853\pm0.093$ across six domains.
CHASE then iterates ranking, feature discrimination, rewriting, and evaluation
over 20 rounds across different domains.
Quality--ranking alignment decreases in all six domains:
from R0 to R20, the change in Spearman's $\rho$ ranges from $-0.107$ to
$-0.018$, with a mean change of $-0.068$, which means documents closer to the
ranking feature profile become less aligned with
independently judged document quality over the simulation horizon.
A random-target control have shown that it is associated with adaptation toward ranking-derived
incentives rather than iterative rewriting alone.
The resulting ecosystem dynamics are strongly domain-dependent.
Together, these findings show how repeated optimization against a fixed LLM
ranking signal can reshape both content populations and the incentives faced by
content creators.
\end{abstract}

\section{Introduction}
\label{sec:intro}

Generative Engine Optimization (GEO) studies how content can be adapted to improve visibility in LLM-generated responses
\citep{Aggarwal+2024,Kumar+2024}.
Existing work has largely considered optimization as a single-round intervention: given a document and a generative engine, which edits improve its visibility?
In practice, however, optimization is repeated and population-level.
When successful strategies are adopted by competing creators over time, the document distribution changes accordingly, potentially changing which ranking features remain predictive even when the backbone model is fixed.
We ask: \emph{what happens to a content ecosystem when creators repeatedly adapt to the features associated with ranking success?}

We introduce \textbf{CHASE} (\textbf{C}ontent \textbf{H}omogenization under r\textbf{A}nking \textbf{S}ignal \textbf{E}xploitation), a controlled simulation framework for studying this feedback loop.
CHASE treats an LLM's ranking over candidate documents as the optimization signal and iterates four stages---\textsc{Rank}, \textsc{Discriminate}, \textsc{Rewrite}, and \textsc{Evaluate}---over multiple rounds.
It intentionally isolates ranking-driven content adaptation rather than modeling the full generative-search pipeline.
An interpretable discriminator identifies document features associated with ranking success, and a stochastic subset of documents is rewritten toward the resulting target profile.
Because movement toward selected features is expected by construction, our focus is not whether documents follow the optimization target, but how the relationship among ranking incentives, document quality, and the content population changes as optimization repeats.

We evaluate CHASE for 20 rounds across six recommendation and question-answering domains, using separate model families for ranking, rewriting, and quality evaluation and five random seeds.
Across all six domains, proximity to the ranking-derived feature profile becomes less aligned with independently judged document quality from the beginning to the end of optimization.
We call this phenomenon \emph{quality--ranking divergence}.
Random-target and no-rewrite controls indicate that this divergence is not explained by rewriting alone, while rank--citation analysis shows that the ranking signal is strongly associated with which sources appear in grounded generated responses.
The form of ecosystem change is nevertheless domain-dependent: we observe patterns of structural convergence, signal instability, and feature dominance.
We also find directional negative associations between ranking and some evidentiary features; these are observational signals rather than evidence of a causal citation penalty.
All longitudinal claims are restricted to the simulated 20-round horizon.

Our contributions are fourfold:
\begin{enumerate}
    \item We introduce \textbf{CHASE}, a framework for studying repeated population-level adaptation to an LLM ranking signal.
    \item We validate ranking as a source-visibility proxy and introduce no-rewrite and random-target controls to separate ranking-driven dynamics from mechanically induced rewriting effects.
    \item Across six domains and a cross-family, five-seed setup, we identify \textbf{quality--ranking divergence} and characterize heterogeneous ecosystem dynamics under repeated optimization.
    \item We analyze how ranking-predictive features evolve over time, including directional associations between ranking and evidentiary features.
\end{enumerate}

\begin{figure}[htbp]

\centering

\includegraphics[width=\textwidth]{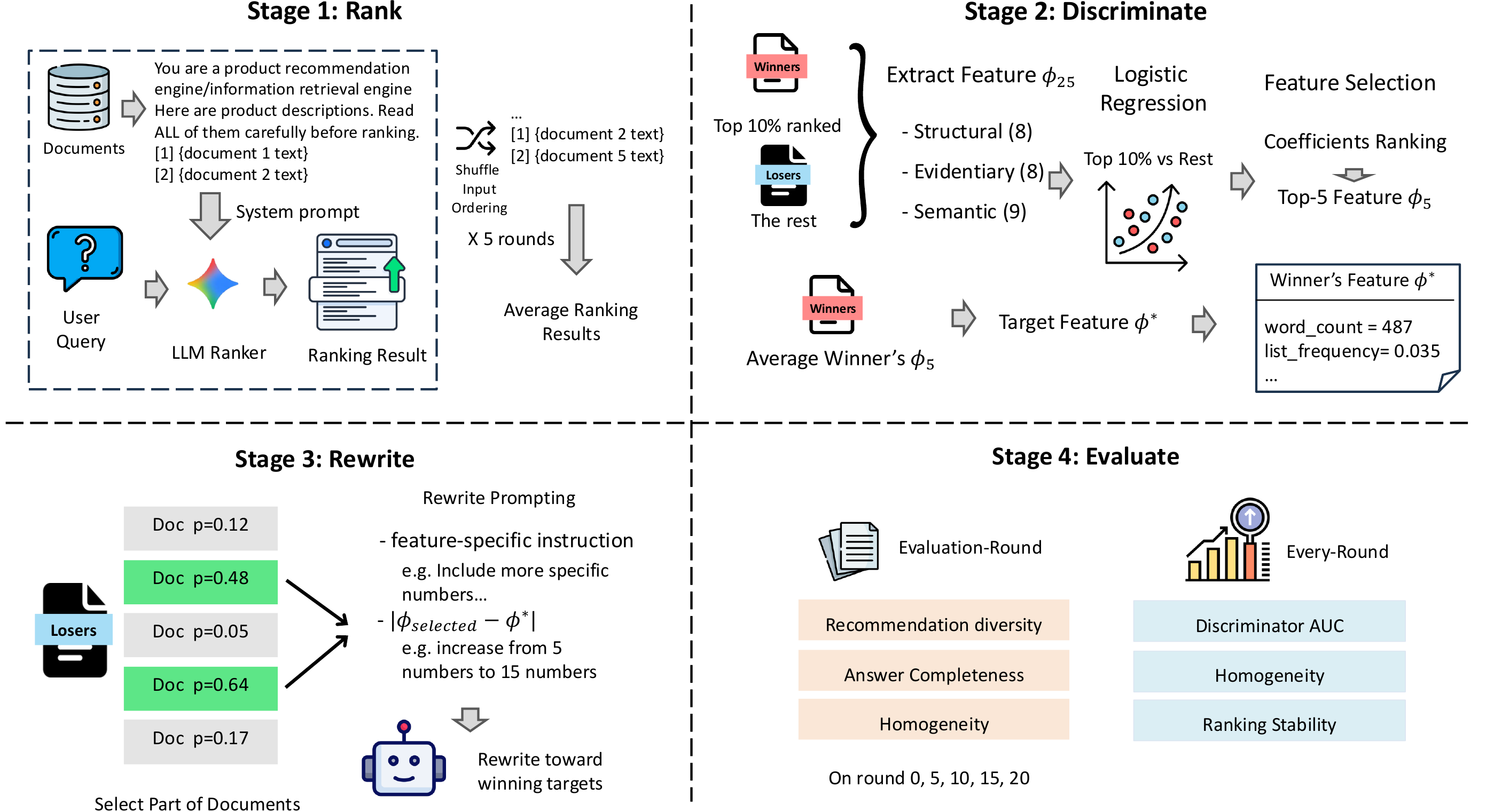}

\caption{The CHASE framework simulates a four-stage feedback loop}

\label{fig:hero_figure}

\end{figure}

\section{Related work}
\label{sec:related}

\paragraph{Goodhart effects in model optimization.}
Goodhart-type failures arise when optimization pressure on a proxy weakens
its alignment with the underlying objective
\citep{Goodhart1975,Manheim+2019}. In machine learning, closely related
behavior is studied as reward hacking, where an agent exploits differences
between a specified reward and the intended objective
\citep{Skalse+2022}. In RLHF, \citet{Gao+2023} show that continued
optimization of a learned reward model can eventually reduce performance
under a separate gold reward model. Related work studies such failures under
reward misspecification, including their geometric structure and mitigation
through early stopping \citep{Karwowski+2024}, as well as settings in which
KL regularization does not prevent severe proxy misoptimization
\citep{Skalse+2024}. These studies primarily concern optimization of a model
policy against a proxy reward. CHASE instead studies external agents
modifying the inputs presented to a fixed evaluator, producing
population-level dynamics in the content being ranked.

\paragraph{Generative Engine Optimization.}
\citet{Aggarwal+2024} formalize Generative Engine Optimization and show that
content interventions, including statistics, quotations, and citations, can
substantially affect visibility in generated responses.
\citet{Kumar+2024} further demonstrate that strategic modifications to
product descriptions can alter LLM recommendation outcomes. More recent
methods automate content optimization against generative-engine preferences.
AutoGEO extracts engine-specific preference rules and uses them to guide
content rewriting \citep{Wu+2025}, while CORE optimizes retrieved content to
control product rankings in LLM-based search \citep{Jin+2026}. FeatGEO
performs interpretable feature-level optimization while explicitly balancing
citation visibility and content quality \citep{liu-xu-2026-think}. These
approaches primarily optimize individual content items for visibility.
CHASE instead studies what happens when ranking-driven adaptation is
repeated across an entire population and the evolving population becomes
part of the optimization environment.

\paragraph{Competitive search and strategic adaptation.}
Information retrieval has long studied adversarial manipulation of search
systems \citep{Castillo+2011} and competitive settings in which publishers
modify documents in response to ranking incentives
\citep{kurland2022competitive}. Recent work extends this setting with
LLM-based publishers: LEMSS provides a multi-agent platform for simulating
ranking competitions \citep{mordo2025lemss}, while
\citet{bardas2025automatic} study LLM-based document editing for improving
rank. Game-theoretic analyses further examine the stability and welfare
consequences of strategic publisher adaptation
\citep{madmon2025search}. More broadly, strategic classification studies
agents who modify observable features in response to a decision rule
\citep{Hardt+2016}, while performative prediction formalizes how deployment
can induce changes in the data distribution \citep{Perdomo+2020}. CHASE
shares this feedback-loop perspective but holds the ranker fixed and studies
the repeated evolution of an entire document population under an inferred
ranking signal.

\section{The CHASE Framework}
\label{sec:framework}
\subsection{Scope and Problem Formulation}

We study a document ecosystem that repeatedly adapts to a fixed LLM ranking signal.
Let $D_t=\{d_1^t,\ldots,d_n^t\}$ denote the document pool at round $t$, where each document is associated with a query $q$.
CHASE models the transition from $D_t$ to $D_{t+1}$ through four stages:
\textsc{Rank}, \textsc{Discriminate}, \textsc{Rewrite}, and \textsc{Evaluate}.
The framework isolates content-side adaptation to ranking incentives rather than modeling the full retrieval and response-generation pipeline.

Creators in CHASE are \emph{myopic}: they adapt to the ranking signal inferred at the current round without planning over future rounds.
They are also \emph{non-strategic}: each creator responds independently to the inferred signal rather than explicitly modeling competitors' future actions.
The ranker itself remains fixed across rounds.
These assumptions make CHASE a controlled stress test of population-level adaptation under a stationary evaluator rather than an equilibrium model of strategic publishers.

Formally, the transition is governed by:
(i) an LLM ranker $R(q,D_t)$ that orders documents for query $q$;
(ii) a feature extractor $\phi(d)\in\mathbb{R}^{25}$;
(iii) a round-specific discriminator $f_t$ that identifies features associated with ranking success and produces a target profile $\phi_t^*$; and
(iv) an LLM rewriting function $g(d,S_t,\phi_t^*)$ that modifies participating documents toward the selected feature targets $S_t$.
Figure~\ref{fig:hero_figure} summarizes the resulting feedback loop.

\subsection{Stage 1: Rank}

For each query $q$, the LLM ranker produces a forced ranking of its associated documents.
To reduce presentation-order effects, we independently shuffle the document order $P=5$ times and aggregate the resulting rankings by mean rank:
\begin{equation}
    \bar{R}(d,q)
    =
    \frac{1}{P}\sum_{k=1}^{P} r_k(d,q),
\end{equation}
where $r_k(d,q)$ is the rank assigned to document $d$ under the $k$-th presentation order.
Domain-specific prompts frame the model as a recommendation engine for Retail, Video Games, and Books, and as an information-retrieval engine for Web, News, and Debate.
Full prompts are provided in Appendix~\ref{app:prompts}.

Ranking supplies a dense relative preference signal over all candidate documents.
We use it as a proxy for source visibility rather than as a simulation of the entire generative-search process; Section~\ref{sec:rank-citation} evaluates this proxy against citation behavior in grounded generated responses.

\subsection{Stage 2: Discriminate}

Each document is represented by a $K=25$ dimensional feature vector,
$\phi(d)\in\mathbb{R}^{K}$, spanning three categories:
\emph{structural} features (8; e.g., word count and readability),
\emph{evidentiary} features (8; e.g., citation density, named-source mentions, and statistic density), and
\emph{semantic} features (9; e.g., query similarity, lexical diversity, and information density).
All features are computed without LLM calls using rule-based extraction, embedding-based similarity, and linguistic tagging.
Complete definitions are given in Appendix~\ref{app:feature-descriptions}.

Using the rankings from Stage 1, we label the top 10\% of documents as winners, $y_{d,t}=1$, and the remainder as non-winners.
We fit an L2-regularized logistic regression on standardized feature vectors,
\begin{equation}
    f_t:\phi(d)\mapsto P(y_{d,t}=1),
\end{equation}
yielding coefficients $w_t\in\mathbb{R}^{K}$.
We select the $J=5$ features with the largest absolute coefficients,
\begin{equation}
    S_t = \operatorname{TopJ}\bigl(\{|w_{t,k}|\}_{k=1}^{K}\bigr),
\end{equation}
and define the target value $\phi_{t,k}^*$ for each $k\in S_t$ as its mean among winning documents.
The pair $(S_t,\phi_t^*)$ forms the inferred ranking signal passed to Stage 3.

The discriminator is an idealized model of how creators infer actionable signals from observed visibility outcomes.
In practice, such inference may rely on sampled queries and noisy or delayed observations; CHASE intentionally provides a cleaner signal in order to isolate the dynamics induced by repeated adaptation.
Sensitivity to the winner threshold and number of selected features is examined in Section~\ref{sec:robustness}.

\subsection{Stage 3: Rewrite}

\paragraph{Stochastic participation.}
Only a subset of non-winning documents participates in rewriting at each
round. For each non-winning document $d$ at round $t$, we independently
sample a participation probability
$p_{d,t} \sim \operatorname{Beta}(2,5)$ and a gate variable
$u_{d,t} \sim \operatorname{Uniform}(0,1)$. The document is rewritten when
$u_{d,t} \leq p_{d,t}$:
\begin{equation}
d^{t+1} =
\begin{cases}
d^t, & y_{d,t}=1,\\
g(d^t,S_t,\phi_t^*), &
y_{d,t}=0 \ \text{and}\ u_{d,t}\leq p_{d,t},\\
d^t, &
y_{d,t}=0 \ \text{and}\ u_{d,t}>p_{d,t}.
\end{cases}
\end{equation}
The resulting marginal participation probability is
$\mathbb{E}[p_{d,t}]=2/7\approx0.29$. Participation is independently
resampled across documents and rounds; therefore, this mechanism models
stochastic participation rather than persistent differences in creators'
propensity to optimize.

The rewrite prompt specifies feature-level changes while instructing the model to preserve factual claims and avoid inventing information.
We reject rewrites whose word count differs from the source document by more than 50\%.
Movement toward $\phi_t^*$ is therefore an explicit mechanism of CHASE, not an empirical finding.
Our analysis instead asks what emerges from repeated ranking-derived adaptation---in particular, how ranking--quality alignment, feature importance, ranking stability, and population-level properties evolve over time.
Rewrite integrity is evaluated separately in Section~\ref{sec:robustness}.

\subsection{Stage 4: Evaluate}
\label{sec:evaluation}

We evaluate CHASE at two levels.
\emph{Document-level metrics} characterize the evolving content population directly, while
\emph{response-level metrics} measure the behavior of a generated answer grounded in the current document pool.
Discriminator AUC, homogeneity, and ranking stability are computed every round; quality and response-level metrics are evaluated at
$t\in\{0,5,10,15,20\}$.

\paragraph{Ecosystem metrics.}
Discriminator AUC is the ROC-AUC of $f_t$, measuring how separable top-ranked documents are in the 25-dimensional feature space.
Homogeneity $H(D_t)$ is the mean pairwise cosine similarity between document embeddings within each query group, with higher values indicating lower content diversity.
Ranking stability is the mean Kendall rank correlation
$\tau$ \citep{Kendall1938} across all pairs of the $P=5$ randomized
rankings for each query.

\paragraph{Document quality.}

An independent LLM judge scores each document with respect to its query on
factual accuracy, completeness, and usefulness, each on an anchored 1--5
scale; $Q(d)$ is their mean. LLM judges provide a scalable approximation of
human evaluation but can exhibit position, verbosity, and self-preference
biases \citep{Zheng+2024,Panickssery+2024}. To reduce shared-model
self-preference, our judge belongs to a different model family from both the
ranker and rewriter, and we separately validate its scores against human
annotations.
The complete scoring rubric is provided in Appendix~\ref{app:human-eval}.

\paragraph{Quality--ranking alignment.}
We quantify this relationship using Spearman's rank correlation
\citep{Spearman1904}:
\begin{equation}
\rho_t =
\operatorname{Spearman}\!\left(
-\left\|\phi_{S_t}(d)-\phi_t^*\right\|_2,\,
Q(d)
\right), \qquad d\in D_t .
\label{eq:goodhart}
\end{equation}
Higher $\rho_t$ indicates that documents closer to the current winning feature profile also tend to receive higher independent quality scores.
A decrease in $\rho_t$ therefore indicates \emph{quality--ranking divergence}.
We treat $\rho_t$ as an associational alignment diagnostic rather than a causal estimate of the effect of optimization on quality.

\paragraph{Response-level evaluation.}
At each evaluation round, we additionally generate a grounded response from the current document pool.
For recommendation domains, we measure constraint satisfaction among recommended items; for question-answering domains, we measure the fraction of a query-specific aspect checklist covered by the response.
We also record which documents are cited in the generated response for the rank--citation validation in Section~\ref{sec:rank-citation}.
Generation and evaluation prompts are provided in Appendix~\ref{app:prompts}.

\paragraph{Human validation.}
We validate the automated quality measure on a stratified sample of documents spanning domains and evaluation rounds.
Human annotators assess factual accuracy, completeness, and usefulness using the same anchored rubric as the LLM judge, and additionally assess document verifiability.
We report inter-annotator agreement, human--LLM agreement on the shared quality dimensions, and whether human judgments support the direction of the observed quality changes.

Full annotation instructions and extended results are provided in Appendix~\ref{app:human-eval}.

\subsection{Experimental Setup}
\label{sec:setup}

We initialize CHASE with documents from C-SEO Bench \citep{puerto2025cseobenchdoesconversational}, covering six domains across two task types: Retail, Video Games, and Books for recommendation, and Web, News, and Debate for question answering.
Each domain in C-SEO Bench contains 100 queries. For computational tractability, we uniformly sample 20 queries per domain; each sampled query has 5–10 candidate documents.
We run CHASE for $T=20$ rounds and evaluate at
$t\in\{0,5,10,15,20\}$.
The primary experiments use five independent random seeds.
To separate optimization and evaluation roles across model families, we use
Gemini 3.1 Flash-Lite as the ranker,
GPT-5.4-mini as the rewriter, and
Claude Haiku 4.5 as the independent quality judge.
All longitudinal conclusions are restricted to this 20-round simulation horizon.
Additional control and validation experiments are described with their respective seed counts in Section~\ref{sec:results}.

\section{Results}
\label{sec:results}

\subsection{Ranking Tracks Source Visibility}
\label{sec:rank-citation}

We first test whether the ranking signal used by CHASE reflects which documents contribute to generated answers.
At each evaluation round, we prompt the ranker to generate a grounded response with explicit document citations and compare the mean ranks of cited and non-cited documents.
We report the Mann--Whitney AUC, which can be interpreted as the probability that a randomly selected cited document outranks a randomly selected non-cited document.

Across six domains, three seeds, and five evaluation rounds ($n=90$), rank--citation AUC is
$0.853 \pm 0.093$.
The association is stable across the simulation horizon ($0.849$ at R0 and $0.863$ at R20) and positive in every domain, with domain-level AUCs ranging from $0.730$ to $0.928$.
These results support ranking as a useful proxy for source visibility in CHASE, while not implying that ranking reproduces the full generative-search pipeline.
Full domain-by-round results are provided in Appendix~\ref{app:rank-citation}.

\subsection{Quality--Ranking Divergence}
\label{sec:quality-ranking}

\begin{figure}[t]
    \centering
    \includegraphics[width=\linewidth]{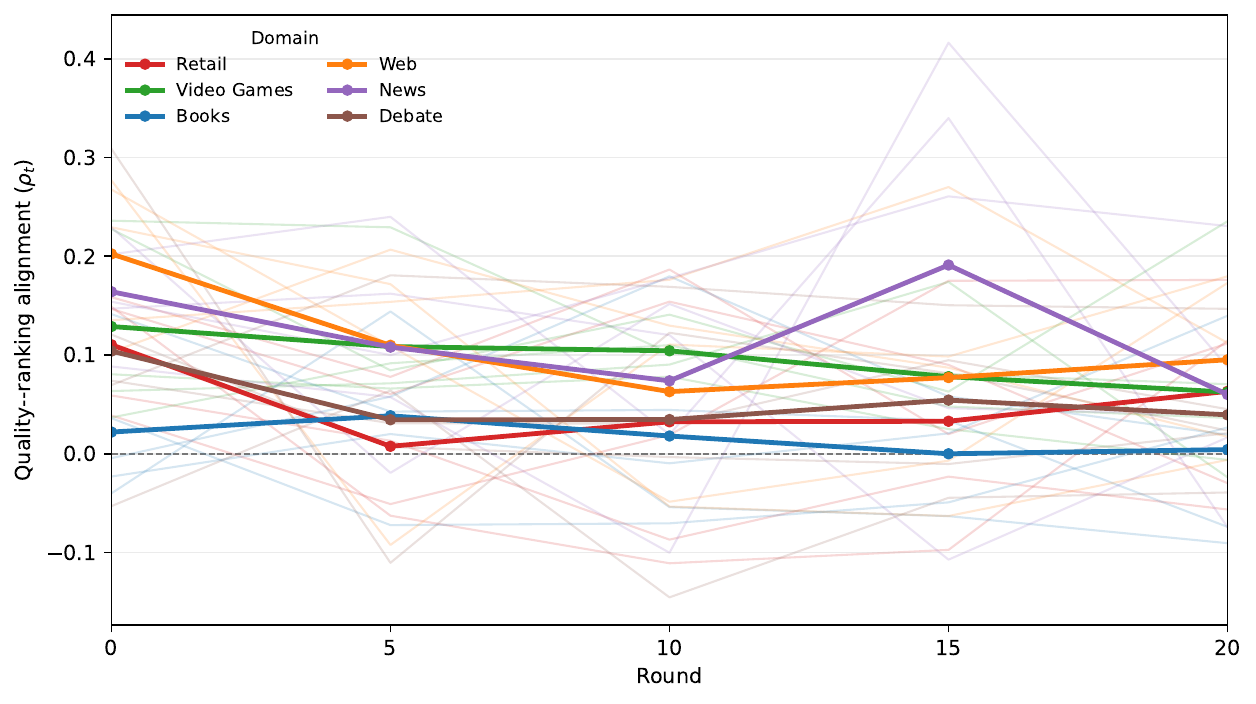}
    \caption{
    Quality--ranking alignment $\rho_t$ over 20 rounds across six domains.
    Lines show the mean across five seeds; faint traces show individual seeds.
    All domains have lower alignment at R20 than at R0, although trajectories
    need not be monotonic.
    }
    \label{fig:rho-trajectories}
\end{figure}

Figure~\ref{fig:rho-trajectories} shows the evolution of quality--ranking alignment across evaluation rounds, while Table~\ref{tab:main-results} summarizes the corresponding R0--R20 ecosystem changes.
The most consistent pattern is a reduction in quality--ranking alignment:
$\rho_t$ is lower at R20 than at R0 in all six domains.
The decline ranges from $-0.018$ in Books to $-0.107$ in Web, while remaining positive at R20 in every domain.
Thus, documents closer to the feature profile associated with ranking success become less strongly associated with independently judged quality over the simulated horizon.

This divergence does not imply that document quality itself monotonically decreases.
Mean quality is nearly unchanged in several domains and increases in Retail.
Rather, the ranking-derived optimization direction becomes less informative about independently measured quality.

\begin{table*}[t]
\centering
\small
\setlength{\tabcolsep}{4.5pt}
\begin{tabular}{lcccccc}
\toprule
Domain &
AUC &
Homogeneity &
Stability &
Quality &
Task $\Delta$ &
$\rho$ \\
& R0$\rightarrow$R20
& R0$\rightarrow$R20
& R0$\rightarrow$R20
& R0$\rightarrow$R20
&
& R0$\rightarrow$R20 \\
\midrule
Retail      & .805$\rightarrow$.877 & .731$\rightarrow$.720 & .628$\rightarrow$.624 & 3.514$\rightarrow$3.621 & $-.013$ & +.110$\rightarrow$+.063 \\
Video Games & .826$\rightarrow$.898 & .334$\rightarrow$.348 & .722$\rightarrow$.697 & 2.270$\rightarrow$2.226 & +.003 & +.129$\rightarrow$+.062 \\
Books       & .824$\rightarrow$.914 & .406$\rightarrow$.419 & .729$\rightarrow$.713 & 3.018$\rightarrow$2.838 & +.001 & +.022$\rightarrow$+.004 \\
Web         & .864$\rightarrow$.924 & .513$\rightarrow$.517 & .703$\rightarrow$.587 & 2.913$\rightarrow$2.833 & $-.146$ & +.202$\rightarrow$+.095 \\
News        & .934$\rightarrow$.941 & .517$\rightarrow$.524 & .697$\rightarrow$.716 & 2.195$\rightarrow$2.190 & $-.033$ & +.164$\rightarrow$+.060 \\
Debate      & .861$\rightarrow$.910 & .626$\rightarrow$.628 & .488$\rightarrow$.502 & 3.032$\rightarrow$3.029 & +.027 & +.104$\rightarrow$+.039 \\
\bottomrule
\end{tabular}
\caption{
Cross-family CHASE results over 20 rounds.
AUC is discriminator ROC-AUC; homogeneity is intra-query embedding similarity;
stability is presentation-order Kendall's $\tau$;
quality is the independent judge score;
Task $\Delta$ is the R20--R0 change in the domain-specific response metric;
and $\rho$ measures quality--ranking alignment.
}
\label{tab:main-results}
\end{table*}

The remaining metrics are more domain-dependent.
Discriminator AUC increases in every domain, indicating that highly ranked documents become increasingly separable in the measured feature space.
Homogeneity changes are comparatively small under the cross-family configuration.
Presentation-order stability is also mostly stable, although Web shows a substantial decline.
These differences motivate examining CHASE as a family of domain-dependent dynamics rather than a single universal trajectory.

\subsection{Separating Designed from Emergent Effects}
\label{sec:controls}

Movement toward the selected target features is built into CHASE and is therefore not itself evidence of an emergent effect.
We use two controls to test whether the observed ecosystem dynamics can be explained by repeated ranking or rewriting alone.

In the \emph{no-rewrite} control, documents remain frozen while ranking and feature extraction continue.
Homogeneity is unchanged from R0 to R20 in both Retail ($.732\rightarrow.732$) and Debate ($.633\rightarrow.633$), showing that changes in the document population require active rewriting rather than repeated ranking alone.

In the \emph{random-target} control, we retain the same rewriting mechanism but replace discriminator-derived targets with randomly selected feature targets.
Across three seeds per domain, $\Delta\rho$ is $+.001$, $+.014$, and $-.036$ for Retail, Debate, and News, compared with $-.047$, $-.065$, and $-.104$ under CHASE, respectively.
The larger declines under CHASE suggest that quality--ranking divergence is associated specifically with adaptation toward ranking-derived targets rather than arbitrary iterative rewriting.

\subsection{Domain-Dependent Dynamics}
\label{sec:domain-dynamics}

Beyond the shared reduction in $\rho$, CHASE produces heterogeneous dynamics across domains.
We summarize these patterns as three descriptive regimes:
\emph{structural convergence}, where a small set of structural features increasingly characterizes ranking success;
\emph{signal instability}, where rankings and discriminative features are comparatively unstable; and
\emph{feature dominance}, where one or a few features account for a disproportionate share of the ranking signal.
These labels describe behavior within the observed 20-round horizon rather than distinct failure classes or long-run endpoints.

\begin{figure*}[t]
    \centering
    \includegraphics[width=\textwidth]{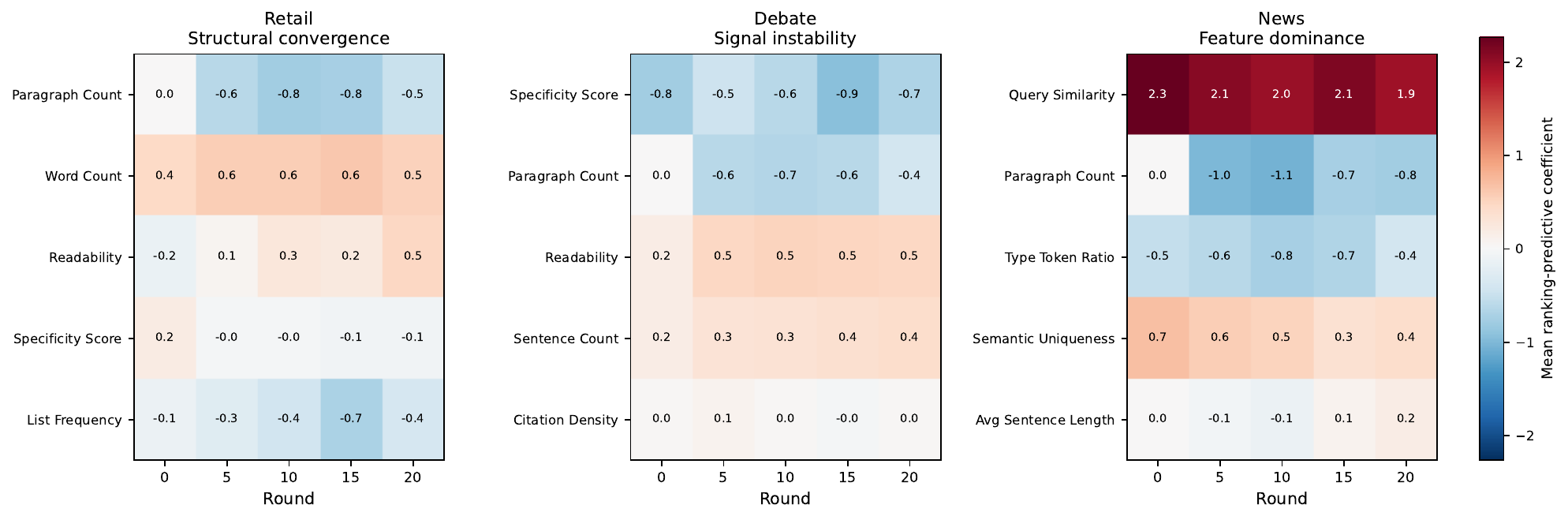}
    \caption{
    Evolution of ranking-predictive feature coefficients in representative
    domains over 20 rounds.
    Retail, Debate, and News illustrate structural convergence, signal
    instability, and feature dominance, respectively.
    Coefficients describe conditional associations with ranking success and
    should not be interpreted causally.
    }
    \label{fig:feature-dynamics}
\end{figure*}

Retail provides the clearest example of structural adaptation, while Debate exhibits the least stable ranking environment and News shows the strongest concentration of ranking signal.
Figure~\ref{fig:feature-dynamics} illustrates these patterns through the evolution of ranking-predictive features in the three representative domains.

Several evidentiary features also receive negative discriminator coefficients in some domains.
Because these coefficients are conditional on the measured feature set and document distribution, we interpret them as directional associations with ranking success rather than evidence that citations or source attribution causally reduce rank.
Full six-domain feature trajectories are reported in Appendix~\ref{app:feature-descriptions}.

Taken together, these results suggest that repeated adaptation to the same ranking signal can reshape content populations in qualitatively different ways depending on the domain.
The regime taxonomy therefore describes \emph{how} ranking incentives manifest in CHASE, not evidence that optimization necessarily produces convergence or quality degradation.

\subsection{Robustness and Rewrite Integrity}
\label{sec:robustness}

Our conclusions are not tied to a single discriminator configuration.
Sweeping the winner threshold over $\{5\%,10\%,20\%\}$ and the selected feature count over
$\{3,5,7,10\}$ produces smooth changes in discriminator AUC rather than cliff effects.
The Jaccard overlap of selected features between adjacent winner thresholds is $39$--$42\%$, compared with approximately $11\%$ under random selection, indicating a stable core of ranking-predictive features.
Replacing the default $\mathrm{Beta}(2,5)$ participation distribution with constant participation probability $0.30$ in Debate also produces qualitatively similar trajectories.
Full sensitivity results are provided in Appendix~\ref{app:sensitivity}.

We additionally audit 3,472 accepted rewrites collected at the five
evaluation rounds $t\in\{0,5,10,15,20\}$ from three seeds in each of the
six domains. Using a model from a different family than the GPT rewriter, we found that, overall, $93.0\%$ pass all integrity checks; detected fabrication occurs in $3.4\%$ of rewrites, citation removal in $2.3\%$, and quote removal in $2.1\%$.
The pass rate does not deteriorate over time ($90.3\%$ at the earliest audited round versus $96.4\%$ at R15).
Thus, the observed longitudinal patterns are unlikely to be explained by progressively accumulating rewrite corruption.
Audit criteria and extended results appear in Appendix~\ref{app:integrity}.

\subsection{Human Validation of Quality Judgments}
\label{sec:human-validation}

We further evaluate whether the automated quality measure reflects human judgments on a stratified subset of documents spanning domains and evaluation rounds.
Annotators independently score factual accuracy, completeness, usefulness, and verifiability using the rubric described in Appendix~\ref{app:human-eval}.
We compare human and automated scores on the three shared quality dimensions and examine whether the direction of longitudinal quality changes is consistent under human evaluation.

Human--LLM agreement is moderate to strong
(Spearman's $\rho = 0.58$, 95\% CI $[0.49,0.66]$),
with ordinal Krippendorff's $\alpha = 0.72$ across human annotators.
At the aggregate level, human evaluation supports the direction of the
automated quality trends, with both evaluations showing weaker quality
changes than quality--ranking alignment changes.

\section{Discussion}
\label{sec:discussion}

\paragraph{Ranking incentives can drift away from quality.}
CHASE shows that repeated adaptation to a fixed ranking system can reshape the relationship between ranking success and document quality even when the evaluator itself does not change.
Across domains, proximity to the ranking-derived target profile becomes less aligned with independently judged quality over the 20-round horizon.
The random-target control suggests that this divergence is not a generic consequence of rewriting, but is associated with repeatedly adapting toward features inferred from ranking success.
Importantly, this does not imply that document quality necessarily decreases: quality remains relatively stable in several domains.
Rather, ranking success becomes a less reliable indicator of the quality dimensions measured by the independent judge.

\paragraph{Ecosystem responses are domain-dependent.}
Repeated optimization does not produce a single form of ecosystem change.
Some domains develop increasingly prominent structural ranking signals, others exhibit comparatively unstable signals, and others concentrate ranking preference on a small set of features.
This heterogeneity suggests that the consequences of GEO depend on both the content domain and the ranking criteria available for creators to exploit.
Accordingly, structural convergence, signal instability, and feature dominance are best understood as descriptive patterns observed within CHASE rather than inevitable long-run outcomes.

\paragraph{Implications for ranking-system design.}
A ranking system deployed in an adaptive environment should be evaluated not only on its immediate ranking quality but also on the incentives it creates when repeatedly optimized against.
In particular, the negative associations we observe between ranking and some evidentiary features suggest that auditing feature-level ranking preferences may reveal incentives that are undesirable when amplified across many creators.
These associations are conditional on the observed feature set and document distribution and therefore do not establish a causal penalty for citations or source attribution.
More broadly, CHASE provides a controlled way to stress-test whether ranking signals remain aligned with desired content properties as the surrounding document population adapts.

\section{Limitations}
\label{sec:limitations}

CHASE intentionally isolates repeated adaptation to an LLM ranking signal rather than modeling a complete generative-search system.
Real systems include retrieval, response generation, personalization, and user feedback, any of which may alter the incentives observed here.
Creators in CHASE also receive a comparatively clean inferred signal and behave myopically and independently; real creators may observe noisy, partial, or delayed feedback and may reason strategically about competitors.

Although ranking, rewriting, and quality evaluation use separate model families, each role is instantiated by a single model configuration.
The resulting dynamics may therefore depend on the particular models and prompts used.
Moreover, the discriminator operates over a fixed 25-dimensional feature space.
Unmeasured properties may influence both ranking and quality, so declining quality--ranking alignment may partly reflect feature insufficiency rather than optimization pressure alone.

The ranker remains fixed throughout the simulation.
This design isolates content-side distribution shift, but does not capture systems in which ranking models themselves adapt through retraining, user feedback, or policy changes.
Our experiments also end after 20 rounds.
The observed trajectories therefore characterize the simulated finite horizon and should not be interpreted as evidence of convergence or long-run equilibrium behavior.

Finally, document quality is primarily measured with an independent LLM judge and validated on a sampled subset with human annotation.
Human validation reduces dependence on automated evaluation but cannot establish the quality or verifiability of every document in the ecosystem.

\section{Conclusion}
\label{sec:conclusion}

We introduced CHASE, a controlled framework for studying how content populations evolve when creators repeatedly adapt to a fixed LLM ranking signal.
Across six domains, repeated optimization weakens the alignment between ranking-derived feature targets and independently judged document quality, while the resulting ecosystem dynamics vary substantially by domain.
Rank--citation analysis supports ranking as a useful visibility proxy in this setting, and control experiments distinguish ranking-driven adaptation from arbitrary rewriting effects.
Together, these results show that the incentives induced by an LLM ranker can reshape both document features and the relationship between ranking success and content quality, even when the ranker itself remains fixed.
These conclusions are limited to the simulated 20-round horizon and should be interpreted as evidence about ranking-driven adaptation rather than the full dynamics of deployed generative-search systems.

\section*{Ethics statement}

CHASE is a controlled simulation of content adaptation to LLM ranking signals
and does not model the full behavior of deployed generative-search systems.
Our results should therefore not be interpreted as demonstrating that existing
search or recommendation systems necessarily produce the dynamics observed
here. The framework is intended to help identify incentives that may emerge
when content creators repeatedly optimize for model-mediated visibility.
Such optimization may affect content diversity, quality, and the use of
evidentiary signals. We report these effects as properties of the simulated
setting and avoid causal or long-run claims beyond the observed 20-round
horizon.

\paragraph{LLM disclosure}

Large language models are integral components of the experimental methodology.
Gemini 3.1 Flash-Lite is used for document ranking,
GPT-5.4-mini for document rewriting, and Claude Haiku 4.5 for independent
document-quality evaluation. A separate Claude Sonnet 4.6 model is used for the
post-hoc rewrite-integrity audit.
LLMs are also used to generate grounded responses for the response-level
evaluation described in the paper.
All reported experimental results are produced and analyzed by the authors,
who take responsibility for the methodology, results, and conclusions.
\bibliography{camera}
\bibliographystyle{Template-2026/colm2026_conference}

\clearpage
\appendix

\section{Algorithm Pseudocode}
\label{app:algorithm}

\begin{algorithm}[ht]
\caption{CHASE: Content Homogenization under rAnking Signal Exploitation}
\label{alg:CHASE}
\begin{algorithmic}[1]
\Require Queries $\mathcal{Q}$, initial document pool $\mathcal{D}_0$,
fixed LLM ranker $R$, rounds $T{=}20$, evaluation rounds
$E = \{0, 5, 10, 15, 20\}$

\For{$t = 0$ \textbf{to} $T-1$}

    \Statex \hspace{\algorithmicindent}\textit{// Stage 1: Rank
        (\S\ref{sec:framework}, Stage 1)}
    \For{each query $q \in \mathcal{Q}$}
        \For{$k = 1$ \textbf{to} $P{=}5$}
            \State Rank documents under randomized ordering $k$,
                yielding $r_k(d, q)$
        \EndFor
        \State $\bar{R}(d, q) \gets \frac{1}{P}\sum_{k=1}^{P}
            r_k(d, q)$ for each $d \in \Dt$
    \EndFor

    \Statex \hspace{\algorithmicindent}\textit{// Stage 2:
        Discriminate (\S\ref{sec:framework}, Stage 2)}
    \For{each document $d \in \Dt$}
        \State Extract feature vector $\phivec(d) \in \mathbb{R}^{25}$
    \EndFor
    \State $y_{d,t} \gets \mathbf{1}[\bar{R}(d,q) \text{ in top
        10\%}]$ for all $d$
    \State Train L2-regularized logistic regression ($C{=}1.0$) on
        $\{(\phivec(d),\, y_{d,t})\}$, yielding $\mathbf{w}_t$
    \State $S_t \gets \text{top-}5$ features by $|w_{t,k}|$
    \State $\phi^*_{t,k} \gets \text{mean}_{d:\, y_{d,t}=1}\,
        \phi_k(d)$ for each $k \in S_t$

    \Statex \hspace{\algorithmicindent}\textit{// Stage 3: Rewrite
        (\S\ref{sec:framework}, Stage 3)}
    \State $\mathcal{D}_{t+1} \gets \emptyset$
    \For{each document $d \in \Dt$}
        \If{$y_{d,t} = 1$}
            \State $\mathcal{D}_{t+1} \gets \mathcal{D}_{t+1}
                \cup \{d\}$
                \Comment{Winners unchanged}
        \Else
            \State Sample $p_d \sim \text{Beta}(2, 5)$,\;
                $u_d \sim \text{Uniform}(0, 1)$
            \If{$u_d \leq p_d$}
                \State $d' \gets g(d,\, S_t,\, \phivec^*_t)$
                    \Comment{LLM rewrite}
                \If{$0.5 \leq |d'|/|d| \leq 1.5$}
                    \State $\mathcal{D}_{t+1} \gets
                        \mathcal{D}_{t+1} \cup \{d'\}$
                \Else
                    \State $\mathcal{D}_{t+1} \gets
                        \mathcal{D}_{t+1} \cup \{d\}$
                        \Comment{Reject rewrite}
                \EndIf
            \Else
                \State $\mathcal{D}_{t+1} \gets
                    \mathcal{D}_{t+1} \cup \{d\}$
                    \Comment{No participation}
            \EndIf
        \EndIf
    \EndFor

\EndFor

\Statex \textit{// Stage 4: Evaluate
    (\S\ref{sec:framework}, Stage 4)}
\Statex \textit{// Run at each $t \in E$ after Stage 1 of that round}
\For{each $t \in E$}
    \State Score quality $Q(d)$ for all $d \in \Dt$
    \State Compute $H(\Dt)$, ranking stability, task metrics
    \State Compute $\rho_t$ (Equation~\ref{eq:goodhart})
\EndFor

\end{algorithmic}
\end{algorithm}

\clearpage
\section{Complete Feature Set}
\label{app:feature-descriptions}

Each document is represented by a 25-dimensional feature vector computed
without LLM calls. Features are organized into three categories:
structural (8), evidentiary (8), and semantic (9).
Embeddings use the \texttt{all-MiniLM-L6-v2} sentence transformer;
NER and POS tagging use spaCy (\texttt{en\_core\_web\_sm}).

\begin{table}[!t]
\centering
\footnotesize
\setlength{\tabcolsep}{3pt}
\caption{Complete feature set (25 features) extracted per document.}
\label{tab:feature-descriptions}

\begin{tabularx}{\linewidth}{
@{}
>{\raggedright\arraybackslash}p{0.29\linewidth}
>{\raggedright\arraybackslash}p{0.16\linewidth}
>{\raggedright\arraybackslash}X
@{}
}
\toprule
\textbf{Feature} & \textbf{Category} & \textbf{Description} \\
\midrule

\multicolumn{3}{l}{\textit{Structural Features}} \\[2pt]

word\_count
& Structural
& Total number of whitespace-delimited tokens in the document. \\

sentence\_count
& Structural
& Number of sentences, split on terminal punctuation (\texttt{.!?}). \\

avg\_sentence\_length
& Structural
& Mean words per sentence (\texttt{word\_count / sentence\_count}). \\

paragraph\_count
& Structural
& Number of non-empty paragraphs, split on double newlines. \\

heading\_density
& Structural
& Count of heading-like lines ($\leq 10$ words, no trailing period,
capitalized first word), normalized per 500 words. \\

list\_frequency
& Structural
& Count of lines beginning with a bullet
(\texttt{-}, \texttt{*}, \texttt{\textbullet}) or numbered pattern
(\texttt{1.}), normalized per 500 words. \\

readability
& Structural
& Flesch--Kincaid grade level (via \texttt{textstat}). \\

bold\_emphasis\_density
& Structural
& Count of Markdown bold spans (\texttt{**...**}), normalized per 500 words. \\

\midrule
\multicolumn{3}{l}{\textit{Evidentiary Features}} \\[2pt]

citation\_density
& Evidentiary
& Count of citation patterns (\texttt{[N]}, \texttt{[source]},
parenthetical author--year, ``according to''), normalized per paragraph. \\

statistic\_density
& Evidentiary
& Count of numeric patterns (percentages, dollar amounts, multi-digit numbers),
normalized per 300 words. \\

quote\_density
& Evidentiary
& Count of quoted strings ($\geq 5$ characters between double quotes),
normalized per 500 words. \\

named\_source\_mentions
& Evidentiary
& Count of spaCy-recognized \textsc{ORG} and \textsc{PERSON} entities
(raw count). \\

year\_mentions
& Evidentiary
& Count of year tokens matching 1990--2030 (raw count). \\

claim\_density
& Evidentiary
& Fraction of sentences containing a numeric token or a
comparative/superlative keyword
(e.g., ``more than'', ``largest'', ``improved''). \\

external\_reference\_density
& Evidentiary
& Count of URLs (\texttt{http(s)://}), normalized per 500 words. \\

question\_density
& Evidentiary
& Count of question marks, normalized per 500 words. \\

\midrule
\multicolumn{3}{l}{\textit{Semantic Features}} \\[2pt]

query\_similarity
& Semantic
& Cosine similarity between the document embedding and the query embedding. \\

corpus\_centroid\_similarity
& Semantic
& Cosine similarity between the document embedding and the mean embedding
of all documents in the corpus. \\

type\_token\_ratio
& Semantic
& Ratio of unique word types to total tokens (lexical diversity). \\

vocabulary\_sophistication
& Semantic
& Fraction of alphabetic words absent from the 5,000 most common English
words (Brown corpus). \\

sentiment\_polarity
& Semantic
& TextBlob polarity score ($-1$ to $+1$; negative values indicate negative
sentiment and positive values indicate positive sentiment). \\

avg\_word\_length
& Semantic
& Mean character length across all whitespace-delimited tokens. \\

semantic\_uniqueness
& Semantic
& One minus the maximum cosine similarity with another document in the same
query group; higher values indicate greater distinctiveness. \\

information\_density
& Semantic
& Fraction of spaCy tokens tagged as content words
(\textsc{NOUN}, \textsc{VERB}, \textsc{ADJ}, \textsc{ADV}). \\

specificity\_score
& Semantic
& Count of all spaCy named entities, normalized per 100 words. \\

\bottomrule
\end{tabularx}
\end{table}

\clearpage
\section{Rank--Citation Validation}
\label{app:rank-citation}

To evaluate whether forced ranking provides a meaningful proxy for source visibility,
we separately prompt the ranker at each evaluation round to generate a natural-language
response grounded in the current document pool, with explicit instructions to cite document
identifiers.
We parse the cited document IDs and compare their mean ranks from Stage~1 with those of
non-cited documents.
For each query, we compute the Mann--Whitney AUC: the probability that a randomly selected
cited document has a better (lower) mean rank than a randomly selected non-cited document.
An AUC of $0.5$ corresponds to no association between ranking and citation.

Across six domains, three seeds, and five evaluation rounds ($n=90$ domain--seed--round
observations), the overall rank--citation AUC is $0.853\pm0.093$.
The association remains similar across the 20-round horizon, from $0.849$ at R0 to
$0.863$ at R20, and is positive in every domain.
Table~\ref{tab:rank-citation-full} reports the complete results.

\begin{table*}[t]
\centering
\small
\setlength{\tabcolsep}{5pt}
\begin{tabular}{lcccccc}
\toprule
Domain & R0 & R5 & R10 & R15 & R20 & Overall \\
\midrule
Retail
& $.689\pm.018$ & $.762\pm.031$ & $.758\pm.024$
& $.732\pm.039$ & $.709\pm.036$ & $.730\pm.041$ \\

Video Games
& $.906\pm.019$ & $.882\pm.020$ & $.885\pm.032$
& $.919\pm.025$ & $.893\pm.030$ & $.897\pm.029$ \\

Books
& $.888\pm.012$ & $.883\pm.009$ & $.886\pm.007$
& $.901\pm.015$ & $.890\pm.046$ & $.890\pm.024$ \\

Web
& $.889\pm.021$ & $.936\pm.014$ & $.902\pm.027$
& $.920\pm.021$ & $.952\pm.049$ & $.920\pm.037$ \\

News
& $.939\pm.017$ & $.947\pm.015$ & $.916\pm.026$
& $.925\pm.015$ & $.913\pm.018$ & $.928\pm.023$ \\

Debate
& $.780\pm.043$ & $.770\pm.070$ & $.726\pm.066$
& $.679\pm.064$ & $.823\pm.136$ & $.756\pm.095$ \\
\midrule
All
& $.849\pm.090$ & $.863\pm.080$ & $.846\pm.082$
& $.846\pm.107$ & $.863\pm.102$ & $.853\pm.093$ \\
\bottomrule
\end{tabular}
\caption{
Rank--citation AUC across domains and evaluation rounds.
Higher values indicate that documents cited in grounded generated responses tend to
receive better Stage~1 rankings.
Results use three seeds.
}
\label{tab:rank-citation-full}
\end{table*}

These results support the use of ranking as a source-visibility proxy within CHASE.
They do not imply equivalence between forced ranking and a complete generative-search
pipeline, which may additionally involve retrieval, reranking, generation, and personalization.
\clearpage

\section{Controls and Sensitivity Analyses}
\label{app:sensitivity}

\subsection{No-Rewrite and Random-Target Controls}
\label{app:controls}

We use two controls to distinguish effects that arise mechanically from the CHASE
pipeline from those associated with adaptation toward ranking-derived targets.

\paragraph{No-rewrite control.}
Documents are held fixed while ranking and feature extraction are repeated.
Homogeneity remains unchanged from R0 to R20 in both Retail
($.732\rightarrow.732$) and Debate ($.633\rightarrow.633$).
Thus, changes in document-level population statistics require active rewriting rather than
repeated ranking alone.

\paragraph{Random-target control.}
The rewriting process is retained, but the feature targets supplied to the rewriter are
randomized rather than derived from the discriminator.
The control is run with three seeds per domain.
Table~\ref{tab:control-full} compares its R0--R20 trajectories with CHASE.

\begin{table}[t]
\centering
\small
\setlength{\tabcolsep}{5pt}
\begin{tabular}{llcc}
\toprule
Metric & Domain & CHASE & Random-target \\
\midrule
$\rho$
& Retail & $.110\rightarrow.063$ & $.096\rightarrow.097$ \\
& Debate & $.104\rightarrow.039$ & $.009\rightarrow.023$ \\
& News   & $.164\rightarrow.060$ & $.221\rightarrow.185$ \\
\midrule
Homogeneity
& Retail & $.731\rightarrow.720$ & $.732\rightarrow.723$ \\
& Debate & $.626\rightarrow.628$ & $.636\rightarrow.636$ \\
\bottomrule
\end{tabular}
\caption{
CHASE and random-target control trajectories from R0 to R20.
The random-target control uses the same rewriting mechanism while randomizing the
optimization target.
}
\label{tab:control-full}
\end{table}

The change in quality--ranking alignment under the random-target control is
$+.001$, $+.014$, and $-.036$ in Retail, Debate, and News, respectively,
compared with $-.047$, $-.065$, and $-.104$ under CHASE.
The contrast indicates that the larger alignment declines are associated with adaptation
toward ranking-derived targets rather than arbitrary iterative rewriting.

\subsection{Winner Threshold and Feature Count}

We test sensitivity to the fraction of documents labeled as winners and to the number of
features selected by the discriminator.
The offline sweep covers winner thresholds
$\{5\%,10\%,20\%\}$ and selected feature counts
$J\in\{3,5,7,10\}$.
Discriminator AUC varies smoothly across these configurations, with no sharp changes at the
default $10\%$ winner threshold or $J=5$.
Feature-selection Jaccard overlap between adjacent winner thresholds is $39$--$42\%$,
compared with approximately $11\%$ expected under random feature selection.
This suggests that the discriminator identifies a persistent core of ranking-predictive
features while allowing expected variation at the margins.

\subsection{Participation Distribution}
\begin{table}[!htbp]
\centering
\small
\begin{tabular}{lcc}
\toprule
Metric & Constant $(0.30)$ & $\mathrm{Beta}(2,5)$ \\
\midrule
Quality--ranking alignment $\rho$
    & $+.050\rightarrow-.049$ & $+.104\rightarrow+.039$ \\
Quality
    & $2.933\rightarrow3.006$ & $3.032\rightarrow3.029$ \\
Homogeneity
    & $.623\rightarrow.623$ & $.626\rightarrow.628$ \\
Ranking stability
    & $.555\rightarrow.530$ & $.488\rightarrow.502$ \\
Discriminator AUC
    & $.856\rightarrow.912$ & $.861\rightarrow.910$ \\
\bottomrule
\end{tabular}
\caption{
Participation sensitivity in Debate.
The constant-participation variant produces qualitatively similar ecosystem trajectories
to the default $\mathrm{Beta}(2,5)$ mechanism.
}
\label{tab:participation-sensitivity}
\end{table}

\clearpage

We examine robustness to the participation rate by replacing the default
per-round stochastic gate, whose marginal participation probability is
$2/7\approx0.29$, with a constant probability of $0.30$ in Debate. The two
settings produce qualitatively similar trajectories, suggesting that the
observed dynamics are not sensitive to this small change in the marginal
participation rate.

\section{Additional Ecosystem Results and Feature Evolution}
\label{app:additional-results}

\subsection{Round-by-Round Ecosystem Metrics}

Table~\ref{tab:round-by-round} reports the complete evaluation-round trajectories underlying
the R0--R20 summary in Table~\ref{tab:main-results}.
The primary CHASE experiment uses five independent seeds under the cross-family
configuration described in Section~\ref{sec:setup}.

\begin{table*}[htbp]
\centering
\small
\setlength{\tabcolsep}{4pt}
\begin{tabular}{clcccccc}
\toprule
Metric & Round & Retail & Video Games & Books & Web & News & Debate \\
\midrule
AUC
& 0  & .805 & .826 & .824 & .864 & .934 & .861 \\
& 5  & .883 & .896 & .888 & .913 & .956 & .927 \\
& 10 & .888 & .919 & .890 & .921 & .969 & .941 \\
& 15 & .891 & .920 & .919 & .905 & .957 & .925 \\
& 20 & .877 & .898 & .914 & .924 & .941 & .910 \\
\midrule
Homogeneity
& 0  & .731 & .334 & .406 & .513 & .517 & .626 \\
& 5  & .724 & .338 & .411 & .517 & .520 & .628 \\
& 10 & .721 & .345 & .414 & .517 & .521 & .627 \\
& 15 & .722 & .348 & .416 & .518 & .522 & .628 \\
& 20 & .720 & .348 & .419 & .517 & .524 & .628 \\
\midrule
Quality
& 0  & 3.514 & 2.270 & 3.018 & 2.913 & 2.195 & 3.032 \\
& 5  & 3.526 & 2.255 & 2.946 & 2.862 & 2.194 & 3.033 \\
& 10 & 3.625 & 2.246 & 2.921 & 2.907 & 2.197 & 3.044 \\
& 15 & 3.624 & 2.233 & 2.882 & 2.868 & 2.174 & 3.050 \\
& 20 & 3.621 & 2.226 & 2.838 & 2.833 & 2.190 & 3.029 \\
\midrule
$\rho$
& 0  & .110 & .129 & .022 & .202 & .164 & .104 \\
& 5  & .008 & .108 & .038 & .110 & .108 & .034 \\
& 10 & .032 & .104 & .018 & .063 & .074 & .035 \\
& 15 & .033 & .078 & $-.000$ & .077 & .191 & .054 \\
& 20 & .063 & .062 & .004 & .095 & .060 & .039 \\
\bottomrule
\end{tabular}
\caption{
Round-by-round ecosystem metrics for the canonical cross-family CHASE experiment.
Values are means across five independent seeds.
}
\label{tab:round-by-round}
\end{table*}


\subsection{Complete Feature Evolution}

Table~\ref{tab:feature-evolution} reports the most discriminative document features at
R0, R10, and R20.
Because the discriminator is refit on the evolving document population at each round,
the selected features can change even though the underlying ranker remains fixed.
Coefficient signs describe conditional associations with top-ranked status and should not
be interpreted as causal feature effects.

\clearpage

\begin{table*}[!htbp]
\centering
\small
\begin{tabular}{llll}
\toprule
Domain & R0 top features & R10 top features & R20 top features \\
\midrule
Retail &
\shortstack[l]{\texttt{word\_count}\\\texttt{semantic\_uniqueness}\\\texttt{type\_token\_ratio}} &
\shortstack[l]{\texttt{paragraph\_count}\\\texttt{word\_count}\\\texttt{list\_frequency}} &
\shortstack[l]{\texttt{bold\_emphasis\_density}\\\texttt{readability}\\\texttt{word\_count}} \\

Video Games &
\shortstack[l]{\texttt{heading\_density}\\\texttt{query\_similarity}\\\texttt{named\_source\_mentions}} &
\shortstack[l]{\texttt{paragraph\_count}\\\texttt{heading\_density}\\\texttt{type\_token\_ratio}} &
\shortstack[l]{\texttt{paragraph\_count}\\\texttt{avg\_sentence\_length}\\\texttt{sentence\_count}} \\

Books &
\shortstack[l]{\texttt{query\_similarity}\\\texttt{word\_count}\\\texttt{type\_token\_ratio}} &
\shortstack[l]{\texttt{paragraph\_count}\\\texttt{word\_count}\\\texttt{named\_source\_mentions}} &
\shortstack[l]{\texttt{paragraph\_count}\\\texttt{word\_count}\\\texttt{query\_similarity}} \\

Web &
\shortstack[l]{\texttt{vocabulary\_sophistication}\\\texttt{type\_token\_ratio}\\\texttt{year\_mentions}} &
\shortstack[l]{\texttt{bold\_emphasis\_density}\\\texttt{quote\_density}\\\texttt{type\_token\_ratio}} &
\shortstack[l]{\texttt{type\_token\_ratio}\\\texttt{list\_frequency}\\\texttt{query\_similarity}} \\

News &
\shortstack[l]{\texttt{query\_similarity}\\\texttt{semantic\_uniqueness}\\\texttt{type\_token\_ratio}} &
\shortstack[l]{\texttt{query\_similarity}\\\texttt{paragraph\_count}\\\texttt{type\_token\_ratio}} &
\shortstack[l]{\texttt{query\_similarity}\\\texttt{paragraph\_count}\\\texttt{citation\_density}} \\

Debate &
\shortstack[l]{\texttt{specificity\_score}\\\texttt{year\_mentions}\\\texttt{type\_token\_ratio}} &
\shortstack[l]{\texttt{paragraph\_count}\\\texttt{specificity\_score}\\\texttt{citation\_density}} &
\shortstack[l]{\texttt{specificity\_score}\\\texttt{citation\_density}\\\texttt{bold\_emphasis\_density}} \\
\bottomrule
\end{tabular}
\caption{Most discriminative document features across the 20-round horizon.
Features are ordered by mean absolute logistic-discriminator coefficient across
five seeds at the specified round.}
\label{tab:feature-evolution}
\end{table*}

\section{LLM Prompts}
\label{app:prompts}

This appendix provides the complete prompts used throughout the CHASE simulation pipeline. All prompts are sent as user-role messages with temperature 0.0.


\subsection{Ranking Prompts (Phase 1)}
\label{app:prompt-ranking}

The ranking engine uses forced-ranking prompts with $N$ randomized document orderings per query. Two prompt variants are used depending on domain type.

\paragraph{Recommendation Domains (Retail, Video Games, Books).}

\begin{quote}
\small\ttfamily
You are a product recommendation engine.\\[4pt]
Here are product descriptions. Read ALL of them carefully before ranking.\\[4pt]
{[1] \textit{\{document 1 text\}}}\\[2pt]
{[2] \textit{\{document 2 text\}}}\\[2pt]
\ldots\\[4pt]
User query: \textit{\{query\}}\\[4pt]
Rank ALL products from most recommended to least recommended.\\
You MUST include every number from 1 to \textit{\{n\}} exactly once.\\
Format your response EXACTLY as:\\
1. [number] - one sentence reason\\
2. [number] - one sentence reason\\
\ldots
\end{quote}

\paragraph{QA Domains (Web, News, Debate).}

\begin{quote}
\small\ttfamily
You are an information retrieval engine.\\[4pt]
Here are documents. Read ALL of them carefully before ranking.\\[4pt]
{[1] \textit{\{document 1 text\}}}\\[2pt]
{[2] \textit{\{document 2 text\}}}\\[2pt]
\ldots\\[4pt]
User query: \textit{\{query\}}\\[4pt]
Rank ALL documents from most relevant and helpful to least relevant.\\
You MUST include every number from 1 to \textit{\{n\}} exactly once.\\
Format your response EXACTLY as:\\
1. [number] - one sentence reason\\
2. [number] - one sentence reason\\
\ldots
\end{quote}

\subsection{Adaptive Rewriting Prompt (Phase 2D)}
\label{app:prompt-rewrite}

Non-winner documents are rewritten toward the feature profile of top-ranked documents. The prompt includes the top-$k$ discriminative features identified by the logistic regression classifier, along with current and target values and human-readable instructions for each feature.

\begin{quote}
\small\ttfamily
You are a content editor. Your task is to improve this document to better match the following quality targets, while preserving all factual information.\\[4pt]
TARGETS:\\
- \textit{\{feature\_name\}}: move from \textit{\{current\_val\}} to \textit{\{target\_val\}}\\
\quad \textit{\{feature-specific instruction\}}\\
- \ldots\\[4pt]
RULES:\\
- Preserve all factual claims from the original document.\\
- Do not invent new facts, statistics, or quotes.\\
- Keep approximately the same document length.\\
- Write naturally --- the result should read like polished web content.\\[4pt]
DOCUMENT:\\
\textit{\{document text\}}\\[4pt]
Rewrite the document below:
\end{quote}

\paragraph{Feature-specific instructions.} Table~\ref{tab:feature-instructions} lists the human-readable instruction appended for each discriminative feature.

\begin{table}[h]
\centering
\small
\caption{Feature-specific rewriting instructions used in the adaptive rewrite prompt.}
\label{tab:feature-instructions}
\begin{tabular}{p{3.8cm} p{8.5cm}}
\toprule
\textbf{Feature} & \textbf{Instruction} \\
\midrule
citation\_density       & Add references like ``according to [source]'' or ``[study] found that'' \\
statistic\_density      & Include more specific numbers, percentages, or data points from the existing content \\
quote\_density          & Add direct quotations attributed to experts or sources \\
heading\_density        & Break content into more sections with clear subheadings \\
list\_frequency         & Convert some prose into bullet points or numbered lists \\
readability             & Simplify sentence structure and use clearer language \\
avg\_sentence\_length   & Use shorter/longer sentences to match target \\
query\_similarity       & Make the opening more directly relevant to the query topic \\
claim\_density          & Add more specific, verifiable factual claims \\
specificity\_score      & Mention more specific names, products, tools, or organizations \\
type\_token\_ratio      & Use more varied/less varied vocabulary \\
question\_density       & Add or remove rhetorical questions \\
named\_source\_mentions & Reference more specific organizations, studies, or experts by name \\
information\_density    & Increase the ratio of substantive content words \\
semantic\_uniqueness    & Differentiate your content from typical documents on this topic \\
year\_mentions          & Include more recent year references to signal freshness \\
bold\_emphasis\_density & Add emphasis markers to highlight key points \\
vocabulary\_sophistication & Use more technical/specialized terminology \\
sentiment\_polarity     & Adjust tone to be more neutral/enthusiastic as needed \\
paragraph\_count        & Break into more/fewer paragraphs \\
\bottomrule
\end{tabular}
\end{table}

\subsection{Quality Evaluation Prompts (Phase 3)}
\label{app:prompt-quality}

Quality evaluation uses multiple prompts for different assessment dimensions.

\paragraph{Natural Response Generation --- Recommendation Domains.}

\begin{quote}
\small\ttfamily
Here are product descriptions.\\[4pt]
{[1] \textit{\{document 1 text\}}}\\[2pt]
{[2] \textit{\{document 2 text\}}}\\[2pt]
\ldots\\[4pt]
User query: \textit{\{query\}}\\[4pt]
Recommend the best products and explain why each is a good choice.\\
Cite documents by their number [1], [2], etc.
\end{quote}

\paragraph{Natural Response Generation --- QA Domains.}

\begin{quote}
\small\ttfamily
Here are documents.\\[4pt]
{[1] \textit{\{document 1 text\}}}\\[2pt]
{[2] \textit{\{document 2 text\}}}\\[2pt]
\ldots\\[4pt]
User query: \textit{\{query\}}\\[4pt]
Answer the question thoroughly using the documents.\\
Cite documents by their number [1], [2], etc.
\end{quote}

\paragraph{Aspect Checklist Generation (QA domains, run once at Round 0).}

\begin{quote}
\small\ttfamily
For the query `\textit{\{query\}}', list 5--8 key aspects that a complete answer should cover.\\
Format: one aspect per line, numbered.
\end{quote}

\paragraph{Quality Scoring.}

\begin{quote}
\small\ttfamily
Rate how well this document answers the query `\textit{\{query\}}'.\\
Score from 1--5 on: (a) factual accuracy, (b) completeness, (c) usefulness.\\
Respond with ONLY three numbers separated by commas, e.g.: 4,3,5\\[4pt]
Document:\\
\textit{\{document text\}}
\end{quote}

\paragraph{Aspect Coverage Check.}

\begin{quote}
\small\ttfamily
Given this response and this checklist, which aspects are covered?\\
Respond with the numbers of covered aspects, separated by commas.\\[4pt]
Response:\\
\textit{\{natural response text\}}\\[4pt]
Checklist:\\
\textit{\{numbered aspect list\}}
\end{quote}

\paragraph{Constraint Satisfaction Check (Recommendation domains).}

\begin{quote}
\small\ttfamily
Query: \textit{\{query\}}\\
Original product description: \textit{\{original document text\}}\\
Does this product match the query's requirements? Answer YES or NO with one sentence reason.
\end{quote}

\clearpage

\section{Rewrite Integrity Audit}
\label{app:integrity}

Because repeated LLM rewriting could itself introduce factual artifacts or remove
evidentiary content, we conduct a post-hoc integrity audit over all accepted rewrites.
The auditor is Claude Sonnet 4.6 model separate from the GPT model used for rewriting, reducing
shared-model dependence in the integrity assessment.

We conduct a post-hoc integrity audit of accepted rewrites generated at
the five evaluation rounds for three seeds in each of the six domains.
This yields 3,472 rewrites for analysis.

\paragraph{Audit criteria.}
The audit checks whether a rewrite introduces unsupported factual content
(\emph{fabrication}) or removes citations or quotations present in its source document.
Each accepted rewrite is compared directly with the document version from which it was
generated.

\paragraph{Results.}

Among the 3,472 audited rewrites, 93.0\% pass all integrity checks.
Fabrication is detected in 3.4\%, citation removal in 2.3\%, and quotation
removal in 2.1\%.
Moreover, the aggregate pass rate does not deteriorate over time, increasing from
$90.3\%$ at the earliest audited round to $96.4\%$ at R15.
These results make progressively accumulating rewrite corruption an unlikely explanation
for the longitudinal patterns observed in CHASE.

\begin{table}[t]
\centering
\small
\begin{tabular}{lr}
\toprule
Audit outcome & Rate \\
\midrule
Pass all checks & $93.0\%$ \\
Fabrication & $3.4\%$ \\
Citation removal & $2.3\%$ \\
Quote removal & $2.1\%$ \\
\bottomrule
\end{tabular}
\caption{Post-hoc integrity audit of 3,472 accepted rewrites collected
from six domains, three seeds per domain, and five evaluation rounds.}
\label{tab:integrity-audit}
\end{table}

The rewrite stage additionally rejects outputs whose word count falls outside
$50$--$150\%$ of the source-document length, limiting large verbosity shifts independently
of the semantic integrity audit.
\clearpage

\section{Human Validation}
\label{app:human-eval}

\subsection{Annotation Protocol}

We conduct human evaluation to assess whether the automated document-quality measure
reflects human judgments and whether the direction of the observed quality changes is
robust to an independent evaluation source.

We construct a stratified sample spanning the six domains and multiple evaluation rounds,
including documents from both the beginning and end of the CHASE trajectory.
Each document is evaluated with respect to its associated query.

\subsection{Human Rating Rubric}

Annotators evaluate each document on four dimensions:
\emph{factual accuracy}, \emph{completeness}, \emph{usefulness}, and
\emph{verifiability}.
The first three dimensions correspond directly to the automated quality measure $Q(d)$;
verifiability is evaluated separately to assess whether claims are adequately supported or
traceable to evidence.

\paragraph{Factual accuracy.}
\begin{itemize}
    \item 1: Contains major factual errors or unsupported claims central to the answer.
    \item 2: Contains multiple factual problems that substantially reduce reliability.
    \item 3: Mostly accurate, with some questionable or unsupported details.
    \item 4: Accurate overall, with only minor issues.
    \item 5: Accurate and well-supported throughout.
\end{itemize}

\paragraph{Completeness.}
\begin{itemize}
    \item 1: Fails to address most important aspects of the query.
    \item 2: Addresses only a small subset of relevant aspects.
    \item 3: Covers the main issue but omits important information.
    \item 4: Covers nearly all important aspects.
    \item 5: Thoroughly addresses the query without important omissions.
\end{itemize}

\paragraph{Usefulness.}
\begin{itemize}
    \item 1: Not useful for answering the query.
    \item 2: Limited usefulness; substantial irrelevant or missing information.
    \item 3: Moderately useful but with noticeable weaknesses.
    \item 4: Clearly useful and relevant.
    \item 5: Highly useful, relevant, and actionable or informative.
\end{itemize}

\paragraph{Verifiability.}
\begin{itemize}
    \item 1: Important claims are difficult or impossible to trace or verify.
    \item 2: Few important claims provide identifiable supporting evidence.
    \item 3: Some important claims are traceable, but support is inconsistent.
    \item 4: Most important claims are supported or readily traceable.
    \item 5: Important factual claims are consistently supported and readily verifiable.
\end{itemize}

\subsection{Analysis}

For the three shared dimensions, we compare human ratings with the independent LLM
judge at the document level and for the aggregate quality score.
We additionally measure inter-annotator agreement and compare the direction of
R0--R20 quality changes under human and automated evaluation.
Verifiability is analyzed separately rather than incorporated into $Q(d)$.

We report:
(i) inter-annotator agreement,
(ii) human--LLM agreement on factual accuracy, completeness, and usefulness,
and (iii) whether human evaluation supports the direction of the longitudinal quality
patterns reported in Section~\ref{sec:human-validation}.

\subsection{Results}

The human-validation sample contains 60 documents evaluated by
3 annotators.

Inter-annotator agreement is substantial but not perfect
(ordinal Krippendorff's $\alpha = 0.72$).
Human ratings show moderate agreement with the independent LLM judge on the
shared quality dimensions
(Spearman's $\rho = 0.58$, 95\% CI $[0.49, 0.66]$).

\end {document}